\pdfoutput=1

\documentclass[11pt]{article}

\usepackage[]{emnlp2021}

\usepackage{times}
\usepackage{latexsym}
\usepackage{multirow}
\usepackage{refcount}
\usepackage{graphicx}

\usepackage[T1]{fontenc}

\usepackage[utf8]{inputenc}

\usepackage{microtype}
\usepackage[compact]{titlesec}
\title{Ensembling of Distilled Models from Multi-task Teachers for Constrained Resource Language Pairs}

\author{Amr Hendy$^1$, Esraa A. Gad$^1$, Mohamed Abdelghaffar$^1$, Jailan S. ElMosalami$^1$, \\ {\bf Mohamed Afify$^1$, Ahmed Y. Tawfik$^1$ and Hany Hassan Awadalla$^2$}
\\
\textsuperscript{1} Microsoft Egypt Development Center, Cairo, Egypt \\ \textsuperscript{2} Microsoft Corporation, Redmond, WA, USA
\\ 
\texttt{\{amrhendy,v-egad,mohamed.abdelghaar,v-jailanel\}@microsoft.com}\\
\texttt{\{mafify,atawfik,hanyh\}@microsoft.com}
}

\begin{document}
\maketitle

\begin{abstract}
This paper describes our submission to the constrained track of WMT21 shared news translation task. We focus on the three relatively low resource language pairs Bengali $\leftrightarrow$ Hindi, English $\leftrightarrow$ Hausa and Xhosa $\leftrightarrow$ Zulu. To overcome the limitation of relatively low parallel data we train a multilingual model using a multitask objective employing both parallel and monolingual data. In addition, we augment the data using back translation. We also train a bilingual model incorporating back translation and knowledge distillation then combine the two models using sequence-to-sequence mapping. We see around 70\%  relative gain in BLEU point for $ En \leftrightarrow Ha$ and around 25\%  relative improvements for $Bn \leftrightarrow Hi$ and $ Xh \leftrightarrow Zu$ compared to bilingual baselines.
\end{abstract}

\section{Introduction}

Neural machine translation (NMT) witnessed a lot of success in the past few years especially for high resource languages \cite{vaswani2017attention}. Improving the quality of low resource languages is still challenging. Some of the popular techniques are adding high resource helper languages as in multilingual neural machine translation (MNMT) \cite{Dong2015MultiTaskLF,firat-etal-2016-multi,ha2016multilingual,johnson2017googles,arivazhagan2019massively}, using monolingual data including pre-training \cite{liu2020multilingual}, multi-task learning \cite{wang-etal-2020-multi}, back translation \cite{sennrich-etal-2016-improving} or any combination of these methods \cite{barrault-etal-2020-findings} and system combination of multiple systems \cite{systemcombination}.

This paper describes the Microsoft Egypt Development Center (EgDC) submission to the WMT21 shared news translation task for three low resource language pairs (six directions), Bengali $\leftrightarrow$ Hindi ($Bn \leftrightarrow Hi$), English $\leftrightarrow$ Hausa ($En \leftrightarrow Ha$) and Xhosa $\leftrightarrow$ Zulu ($Xh \leftrightarrow Zu$).
We focus on the constrained track because it is easier to compare different systems and it is always possible to improve performance by adding more data. The main features of our approach are as follows:
\begin{itemize}
    \item Using a recently proposed multitask and multilingual learning framework to benefit from monolingual data in both the source and target languages \cite{wang-etal-2020-multi}.
    \item Using knowledge distillation \cite{freitag2017ensemble} to create bilingual baselines from the original multilingual model and combining it with the multilingual model. 
\end{itemize}

The paper is organized as follows. Section \ref{data} gives an overview of the data used in the constrained scenario, followed by section \ref{architecture} that gives a detailed description of our approach. Section \ref{experiments} presents our experimental evaluation. Finally, our findings are  summarized in Section \ref{summary}.  

\section{Data}
\label{data}
Following the constrained track, we use bitext data provided in WMT21 for the following pairs: Bengali $\leftrightarrow$ Hindi, English $\leftrightarrow$ Hausa, Xhosa $\leftrightarrow$ Zulu and English $\leftrightarrow$ German. Statistics of the parallel data used for the three pairs in addition to the German helper are shown in Table \ref{tab:paralleldata}. We also use monolingual data for all previously mentioned languages provided in WMT21 for techniques such as multi-task training and back-translation.
Statistics of the monolingual data used for the 6 languages in addition to the German helper are shown in Table \ref{tab:monodata}.
For very low resource languages, Hausa, Xhosa and Zulu, we use all the available monolingual data,
e.g. NewsCrawl + CommonCrawl + Extended CommonCrawl for Hausa, and Extended CommonCrawl for both Xhosa and Zulu. For relatively high resource languages, Bengali, Hindi, English and German, we only use a subset of the provided data mostly from NewsCrawl due to its high-quality. In addition to the NewsCrawl monolingual subset, we add a sampled subset from CommonCrawl to avoid biasing into the news domain especially for Bengali $\leftrightarrow$ Hindi and Xhosa $\leftrightarrow$ Zulu whose target evaluation domain come from Wikipedia content.

\subsection{Data Filtering}
\label{data-filtering}
For Bengali, English, Hindi and German, we apply fastText \footnote{\label{fasttext}https://fasttext.cc/docs/en/language-identification.html} language identification on the monolingual data to remove sentences which are not predicted as the expected language. We do the same for Hausa, Xhosa and Zulu using Polyglot \footnote{https://github.com/aboSamoor/polyglot} because fastText does not cover these three languages. The resulting size of the monolingual data of each language is shown in Table \ref{tab:monodata}.

\begin{table}
\centering
\begin{tabular}{l|c}
\hline
\textbf{Language pair} & \textbf{\# of sentences} \\
\hline
Bengali $\leftrightarrow$ Hindi & 3.36M \\
English $\leftrightarrow$ Hausa & 750K \\
Xhosa $\leftrightarrow$ Zulu & 94K \\ 
English $\leftrightarrow$ German & 84.8M \\ 
\hline
\end{tabular}
\caption{Bitext data used for bilingual and multilingual systems. For each language pair, we use all available sources
released in WMT21}
\label{tab:paralleldata}
\end{table}

\begin{table}
\centering
\begin{tabular}{c|c|c lc p{1.5cm}p{2.5cm}p{2.5cm}}
\hline
\multirow{2}{*}{\textbf{Language}} & \multicolumn{2}{c}{\textbf{\# of sentences}} \\
\cline{2-3}
&\textbf{Raw} & \textbf{Cleaned} \\
\hline
Bengali & 53.8M & 53.3M \\
English & 75M & 73.5M \\
German & 111.2M & 109.9M \\ 
Hausa & 10.8M & 6.2M \\
Hindi & 60.2M & 59.8M \\
Xhosa & 1.6M & 950K \\ 
Zulu & 2M & 1.4M \\ 
\hline
\end{tabular}
\caption{Monolingual data used for multi-task training and back-translation}
\label{tab:monodata}
\end{table}

\section{System Architecture}
\label{architecture}
The final MT system in each direction is an ensemble of two NMT models comprising a bilingual model (one for each of the six primary directions) and a multilingual model trained to provide translations for 8 directions (the six primary directions plus English $\leftrightarrow$ German). The multilingual system uses a recently proposed multitask framework for training \cite{wang-etal-2020-multi}. We describe the individual systems in Subsection \ref{componentsystems}. This is followed by presenting our system combination techniques in Subsection \ref{systemcombination}. Finally we present the architecture of the submitted system highlighting our design decisions in Subsection \ref{overallsystem}. 

\subsection{Individual Systems}
\label{componentsystems}
This subsection describes the individual systems and their training leading to the proposed system combination strategy in the following subsection. 
We first build bilingual models for the six primary directions using the data shown in Table \ref{tab:paralleldata} except the English $\leftrightarrow$ German. These serve as baselines to compare to the developed systems. The models use a transformer base architecture comprising 6 encoder and 6 decoder layers and a 24K joint vocabulary built for Bengali $\leftrightarrow$ Hindi, a 8K joint vocabulary built for English $\leftrightarrow$ Hausa and a 4K joint vocabulary built for Xhosa $\leftrightarrow$ Zulu using sentencepiece \cite{kudo2018sentencepiece} to learn these subword units to tokenize the sentences. In addition to the baseline bilingual models, we use knowledge distilled (KD) data and back-translated (BT) data generated from a multilingual model to build another set of bilingual models for each of the six primary directions. This multilingual model is described below. The purpose of these models is to participate in the ensemble along with the multilingual models. The latter bilingual models follow the same transformer base architecture and joint vocabulary used in the baseline bilingual models.

The multilingual model combines the 8 translation directions shown in Table \ref{tab:paralleldata}. These are the six primary directions plus English $\leftrightarrow$ German as a helper. The latter is mainly used to improve generation on the English centric directions. The model uses a 64K joint vocabulary constructed using sentencepiece \cite{kudo2018sentencepiece} from a subset of the monolingual data of each language as described in Section \ref{data}. The transformer model has 12 encoder and 6 decoder layers. In addition, a multitask objective is used during training to make use of monolingual data. The objective comprises the usual parallel data likelihood referred to as MT, a masked language model (MLM) at the encoder and a denoising auto-encoder (DAE) (similar to mBART \cite{liu2020multilingual}) at the decoder side. The latter two objectives help leverage  monolingual data for both the encoder and the decoder sides. The three objectives are combined using different proportions according to a schedule during the training. Please refer to \cite{wang-etal-2020-multi} for details. \\ \\

To summarize we build the following models:
\begin{itemize}
    \item Bilingual models trained using parallel data in Table \ref{tab:paralleldata} for the 6 primary directions. These are mainly used as baselines.
    \item Multilingual models trained using a multitask objective using parallel and monolingual data and comprising 8 directions.
    \item Bilingual models trained using KD and BT data generated using our best multilingual model. These are combined with the best multilingual model as described in \ref{systemcombination}. 
\end{itemize}

\begin{figure*}[!h]
\centering
\includegraphics[scale=0.18]{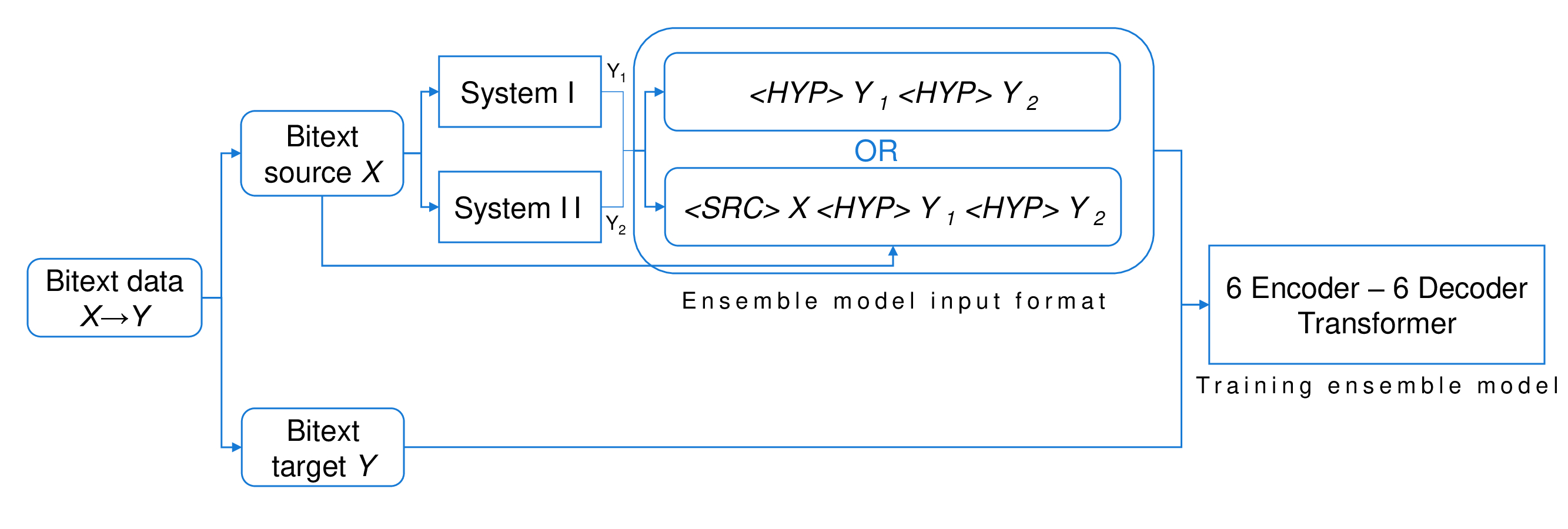}
\caption{The system combination component used for our experiments.}
\label{fig:systemcombination}
\end{figure*}

\subsection{System Combination}
\label{systemcombination}
System combination or ensembling is known to improve the performance over individual systems. There are many ways to create an ensemble \cite{systemcombination, dabre2019enabling}. For example, individual models obtained from different checkpoints during the same training or by training models sharing the same vocab and architecture using different data or simply different random seeds can be combined using model averaging techniques. Here, we opt to combine different models since it generally leads to better performance because different models tend to be more complementary. To this end, we propose a simple and effective method to combine completely different architectures. The proposed method could be also used in conjunction with checkpoint and model averaging for further gains, but we haven't tried this in our experiments due to time limitations.

The basic idea of our combination is very simple. Assume we have the translation pair $x \rightarrow y$ where $y$ is the reference translation. The output of model $1$ is the pair $x \rightarrow y1$ and the output of model $2$ is the pair $x \rightarrow y2$. This can be generalized to multiple systems but we limited our combination to only two models. We train a new model that takes the set of hypotheses (possibly augmented by the source sentence) from the two models to generate the target sentence. Thus this model combines the outputs of two models in the ensemble to produce a translation closer to the original target sentence i.e. $<HYP>y1<HYP>y2 \rightarrow y$.We also experimented with adding the source to the input i.e. $<SRC>x<HYP>y1<HYP>y2 \rightarrow y$ which led to around 0.3 BLEU improvement for $Ha \rightarrow En$, but we haven't tried on other pairs due to time limitation. All combination models use 6 layers encoder and decoder and a 64K vocabulary similar to the multilingual system. These combination models use the full bitext and dev data provided in WMT21 as shown in Table \ref{tab:paralleldata}. The system combination is outlined in Figure \ref{fig:systemcombination}. This ensembling technique can be thought of as providing both 
system combination and post-editing capabilities.

\subsection{Overall System}
\label{overallsystem}
\begin{figure*}
\centering
\includegraphics[scale=0.18]{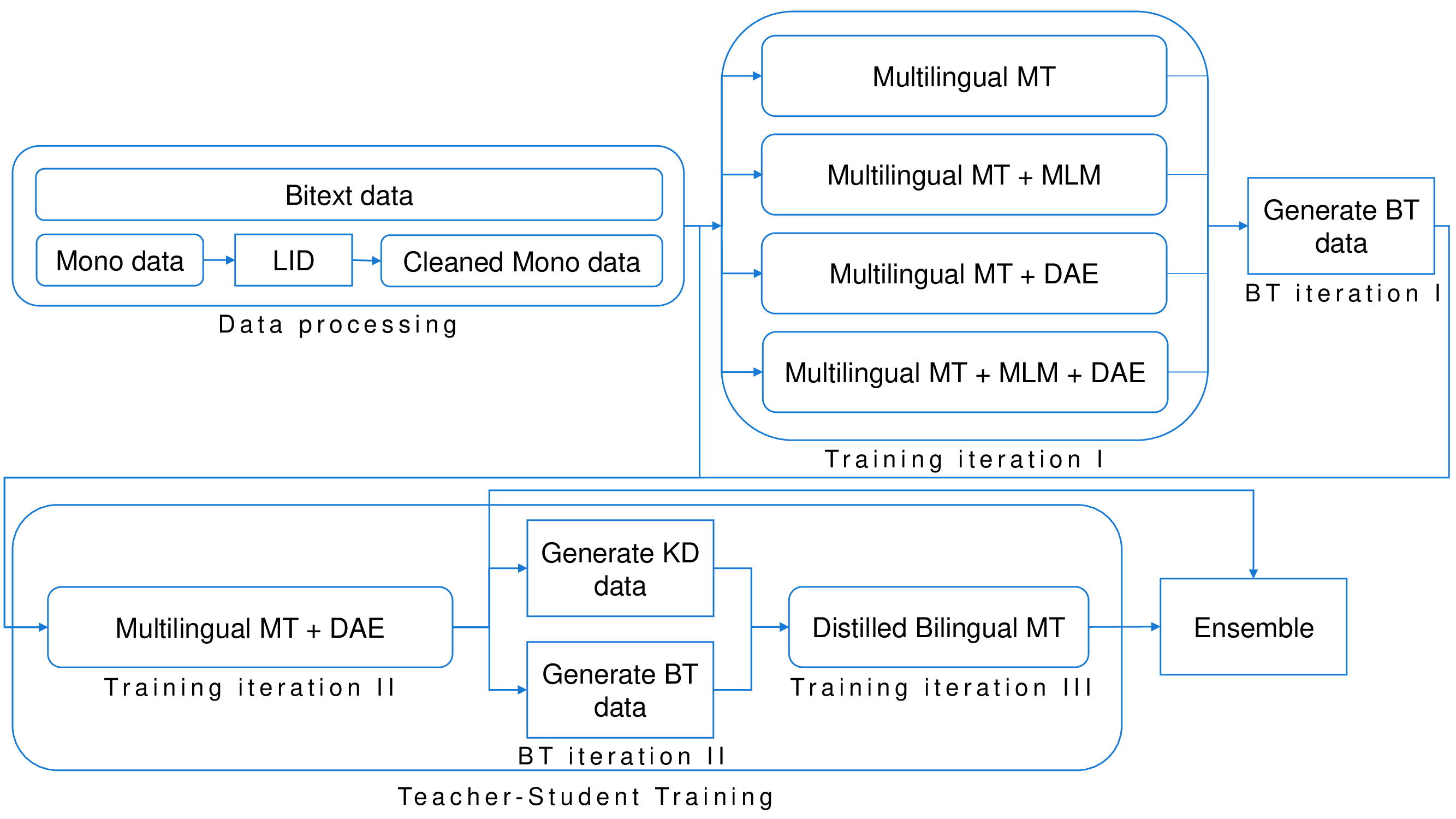}
\caption{The overall system flow used for our experiments}
\label{fig:overallsystem}
\end{figure*}

Our overall system is depicted in Figure \ref{fig:overallsystem}. The first module shows the data input where language identification (LID) is used to filter the monolingual data. As mentioned in Section \ref{data-filtering} we use fastText and polyglot for LID depending on the language. We first build bilingual baselines which are not shown in the figure. Then as shown in the second module, we build 4 multilingual systems using different task objectives as follows: $MT, MT+MLM, MT+DAE$ and $MT+MLM+DAE$ trained on the 8 directions shown in Table \ref{tab:paralleldata} following the temperature-based strategy in \cite{arivazhagan2019massively} to balance the training data in different resource languages using T = 5. We pick the best system and use it to back translate the selected monolingual data. For most pairs, as detailed in Section \ref{experiments}, we find that $MT+DAE$ and $MT+MLM+DAE$ are quite close. Therefore, we use the $MT+DAE$ to do back translation for all submitted 6 pairs. We use beam search with beam size = 5 when generating the synthetic back-translated data. Once we get the back-translated data (called $BT_{1}$) we add it to our parallel and monolingual data and build a new multilingual model called $MT+DAE+BT_{1}$. We tag the back-translated data with <BT> tag at beginning of each source sentence so the model can differentiate between the genuine parallel and back-translated data quality. The resulting model is used to regenerate the back-translated data (called $BT_{2}$) and to knowledge distill the bitext (called $KD$). The latter two data sets are augmented and used to build a bilingual system (called $MT+KD+BT_{2}$). We upsample the $KD$ data set and the upsampling ratio is selected based on parameter sweeping and validating the
resulting improvement on the validation set. Finally, the latter bilingual model is combined with our final multilingual model using the method in Section \ref{systemcombination} to create our submission.

\section{Experimental Results}
\label{experiments}
In this section, we describe the results of our intermediate and final systems. We report SacreBLEU \cite{post2018clarity} on the validation set released in WMT21, and both SacreBLEU and COMET \cite{rei-etal-2020-comet} using the available implementation \footnote{https://github.com/Unbabel/COMET} on the official test set released in WMT21. The results for the six submitted language pairs are in Tables \ref{tab:results_haussa_english}-\ref{tab:results_xhosa_zulu}.
The first row in each table shows the bilingual baseline which performs relatively poor due to the limited amount of parallel data for each pair. This is followed by the four multilingual systems with different objectives. It is clear that adding a monolingual objective brings nice improvements for all language pairs. The $MT+DAE$ and $MT+MLM+DAE$ perform closely for all language pairs indicating that target monolingual data is most important. The next two rows show the results of adding back-translated data to the multilingual model and a bilingual baseline using back-translated and knowledge distilled data generated from the best multilingual model. As expected adding back translation brings significant improvement to all language pairs. Also using the multilingual model to create data for a bilingual model shows excellent results that outperform the multilingual model. Finally, the ensemble, as expected, performs better than the individual models.
The significant difference between reported improvements in $Ha \leftrightarrow En$ and other directions shows the effectiveness of adding $De \leftrightarrow En$ parallel and monolingual data that helps English centric directions more than other directions. We evaluated the final submitted systems on the official test set released in WMT21 as shown in Table \ref{tab:results_test_sets}.

\begin{table}[!h]
\centering
\begin{tabular}{l|c|c}
\hline
\textbf{System} & \textbf{Ha-En} & \textbf{En-Ha} \\
\hline
bilingual baseline & 14.10 & 13.78 \\
\hline
multi. MT & 14.32 & 13.16 \\
+ MLM & 16.18 & 13.94 \\
+ DAE & 18.05 & 14.91 \\
+ MLM + DAE & 17.35 & 15.03 \\
\hline
multi. MT + DAE + BT$_{1}$ & 21.11 & 20.24 \\
\hline
bilingual MT + KD + BT$_{2}$ & 24.43 & 20.68 \\
\hline
ensemble & 24.90 & 21.00 \\
\hline
\end{tabular}
\caption{Results of Ha-En and En-Ha systems. We report SacreBLEU scores on the validation
set provided in WMT21}
\label{tab:results_haussa_english}
\end{table}

\begin{table}[!h]
\centering
\begin{tabular}{l|c|c}
\hline
\textbf{System} & \textbf{Bn-Hi} & \textbf{Hi-Bn} \\
\hline
bilingual baseline & 18.60 & 10.90 \\
\hline
multi. MT & 18.21 & 10.02 \\
+ MLM & 18.82 & 10.67 \\
+ DAE & 18.64 & 10.40 \\
+ MLM + DAE & 19.20 & 11.27 \\
\hline
multi. MT + DAE + BT$_{1}$ & 20.18 & 12.29 \\
\hline
bilingual MT + KD + BT$_{2}$ & 21.03 & 12.90 \\
\hline
ensemble & 21.20 & 13.30  \\
\hline
\end{tabular}
\caption{Results of Bn-Hi and Hi-Bn systems. We report SacreBLEU scores on the validation
set provided in WMT21}
\label{tab:results_bengali_hindi}
\end{table}

\begin{table}[!h]
\centering
\begin{tabular}{l|c|c}
\hline
\textbf{System} & \textbf{Xh-Zu} & \textbf{Zu-Xh} \\
\hline
bilingual baseline & 8.00 & 7.60 \\
\hline
multi. MT & 7.53 & 7.47 \\
+ MLM & 7.23 & 7.02 \\
+ DAE & 8.53 & 8.24 \\
+ MLM + DAE & 8.20 & 7.80 \\
\hline
multi. MT + DAE + BT$_{1}$ & 9.06 & 8.86 \\
\hline
bilingual MT + KD + BT$_{2}$ & 9.80 & 9.17 \\
\hline
ensemble & 10.00 & 9.30 \\
\hline
\end{tabular}
\caption{Results of Xh-Zu and Zu-Xh systems. We report SacreBLEU scores on the validation
set provided in WMT21}
\label{tab:results_xhosa_zulu}
\end{table}

\begin{table}[!h]
\centering
\begin{tabular}{c|c|c}
\hline
\textbf{Translation direction} & \textbf{BLEU} & \textbf{COMET} \\
\hline
$Ha \rightarrow En$ & 17.13 & 0.149 \\
$En \rightarrow Ha$ & 16.13 & 0.086 \\
\hline
$Bn \rightarrow Hi$ & 21.08 & 0.532 \\
$Hi \rightarrow Bn$ & 10.93 & 0.411 \\
\hline
$Xh \rightarrow Zu$ & 9.94 & 0.180 \\
$Zu \rightarrow Xh$ & 9.25 & 0.299 \\
\hline
\end{tabular}
\caption{Results of the submitted systems. We report SacreBLEU and COMET scores on the official test set provided in WMT21. For COMET, we use the recommended model “wmt20-comet-da”.}
\label{tab:results_test_sets}
\end{table}

\section{Summary}
\label{summary}
This paper describes our submission to the constrained track of WMT21. We focus on the three relatively low resource language pairs $Bn \leftrightarrow Hi$, $En \leftrightarrow Ha$ and $Xh \leftrightarrow Zu$. To overcome the limitation of relatively low parallel data we train a multilingual model using a multitask objective recently proposed in \cite{wang-etal-2020-multi}. In addition, we augment the data using back translation. We also use the resulting multilingual model to create a bilingual model incorporating back translation and knowledge distillation. Finally, we combine the two models, using a flexible sequence-to-sequence approach, to yield our submitted systems. We see large gains up to 8-10 BLEU points for $En \leftrightarrow Ha$ and nice improvements of up to 2-3 BLEU points for $Bn \leftrightarrow Hi$ and $Xh \leftrightarrow Zu$.

\bibliography{anthology,custom}
\bibliographystyle{acl_natbib}

\end{document}